\documentclass[twocolumn]{article}
\usepackage{caption} 
\usepackage{multirow}
\usepackage{esvect}
\usepackage{bigfoot}
\usepackage{etoolbox}

\makeatletter
\patchcmd{\maketitle}{\@fnsymbol}{\@alph}{}{}  % Footnote numbers from symbols to small letters
\makeatother
\usepackage{floatrow}
\newfloatcommand{capbtabbox}{table}[][\FBwidth]
\usepackage{tabularx}

\usepackage{blindtext}
\usepackage{geometry}                % See geometry.pdf to learn the layout options. There are lots.
\usepackage{amsmath}

\usepackage{hyperref}

\usepackage{verbatim}

\hypersetup{
    colorlinks=true,
    linkcolor=blue,
    filecolor=magenta,      
    urlcolor=blue,
}
\usepackage{multicol}
\usepackage{mathptmx}       % selects Times Roman as basic font
\usepackage{helvet}         % selects Helvetica as sans-serif font
\usepackage{courier}        % selects Courier as typewriter font
\usepackage{type1cm}        % activate if the above 3 fonts are
\usepackage{enumitem}

\usepackage{makeidx}         % allows index generation
\usepackage{graphicx}        % standard LaTeX graphics tool
\usepackage{multicol}        % used for the two-column index
\usepackage[bottom]{footmisc}% places footnotes at page bottom

\makeindex             % used for the subject index

\title{Informational Neurobayesian Approach to Neural Networks Training. Opportunities and Prospects 
}

\makeatletter
\renewcommand\@date{{%
  \vspace{-\baselineskip}%
  \large\centering
  \begin{tabular}{@{}c@{}}
    Artemov A.\textsuperscript{1} \\
    \normalsize science@cogsys.company
  \end{tabular}%
  \quad  \quad
  \begin{tabular}{@{}c@{}}
    Lutsenko E.\textsuperscript{2} \\
    \normalsize prof.lutsenko@gmail.com
  \end{tabular}

  \begin{tabular}{@{}c@{}}
    Ayunts E.\textsuperscript{3} \\
    \normalsize eayunts@gmail.com
  \end{tabular}
  \begin{tabular}{@{}c@{}}
             Bolokhov I.\textsuperscript{1} \\
    \normalsize ibolokhov@cogsys.company
  \end{tabular}

  \bigskip

  \textsuperscript{1} Cognitive Systems Company, Brain2 project \footnote{Created with the support of the Foundation for Assistance to Small Innovative Enterprises, Russia} \par
  \textsuperscript{2}Kuban State University

  \textsuperscript{3}Higher School of Economics

  \bigskip

}}
\makeatother

\begin{document}
%\begin{multicols}{2}

%\author{A.Artemov, Engineer\footnote{Cognitive Systems,sNeurobayesiancience@cogsys.company} , A. Sergeev\footnote{Cognitive Systems,science@cogsys.company}, I. Khasenevich\footnote{Cognitive Systems,science@cogsys.company} }

%\author{A.Artemov \thanks{Brain2, science@cogsys.companNeurobayesiany} , Lutsenko E.V.  \thanks{Kuban State University, Krasnodar, prof.lutsenko@gmail.com}}
%\date{\vspace{1ex}}

\maketitle

\abstract{A study of the classification problem in context of information theory is presented in the paper.  Current research in that field is focused on optimisation and bayesian approach. Although that gives satisfying results, they require a vast  amount of data and computations to train on. Authors propose a new concept, named Informational Neurobayesian Approach (INA), which allows to solve the same problems, but requires significantly less training data as well as computational power. Experiments were conducted to compare its performance with the traditional one's  and the results showed that capacity of the INA is quite promising. }

\section{Introduction}
The classification problem nowadays is usually solved by complex analytical algorithms requiring plenty of data. Another field of research is neurobayesian approach, which proposes optimisation based on Bayes rule with some tricks to cope with functions complexity and make them well-defined. The authors introduce a method based on the same concepts but they also take into consideration the informational side of the data - namely its meaning. %There are two approaches :  the Shennon one suggests to analyse input data to detect message, while Hartley does the same for the message to classify inputs. In the current paper the 

The problem of determining the quantity of information until recently was based on Hartley and Shennon approaches. The first one represents just the quantity of information, the second is its generalisation which takes into account different probabilities for classes. Another fundamental concept is Harkevich   formula which measures the informational value of event or message as the logarithm of ratio of probability of a class after and before the event. The System Theory of Information (STI) proposed by Lucenko \footnote{ Section 4.2 author,\cite{luc} [Lutsenko E., 2002]} merges these concepts, taking each formula with corresponding power, which illustrates the emergence of classes and features. The main idea of the article is to develop a method, that allows to calculate weight coefficients of artificial neural network’s synapses. We propose to move from uninterpreted weights (trial and error method) to the weight representing the amount of information that is contained in the feature regarding a possible consequence. Thus, a neural network is considered as a self-training system of transforming the input data into the amount of information about the possible consequences.The proposed approach allows to train big neuromodels with billions of neurons on ordinary CPUs.
 
\section{Classical Gradient Descent}
\subsection{Description}

Gradient Descent is a widely used approach to solve optimisation problems which are too complex to find solution analytically. 
As the approach proposed in the paper is applied in the context of neural networks, we will briefly show the usage of gradient descent for their training. 

The problem is formulated in the following way: model's performance is  measured by the error function, which depends on  $n$ neurons and $m$ observations. So, $n$ weights are to be optimised to achieve the minimum of error function.
The gradient descent principle is to push the weights towards the direction opposite to the error function's gradient (all weights are changed simultaneously).
\subsection{Formulae}
\[
\nabla Q(w_0, w_1, ..., w_n) =\Bigg[  \frac {\delta Q}{\delta w_0}, ..., \frac {\delta Q}{\delta w_n}  \Bigg]  \]
So, on each iteration the weights are updated according the following rule:
$$ w_i = w_i - \mu \frac {\delta Q}{\delta w_i},$$
where  $\mu$ is an exogenous parameter, which determines the learning rate. The most popular type of error functions is sum of squares, so the aim is to minimise the expression:
$$ Q(w_0, w_1, ..., w_n) = \frac 12 \sum^m_{j =1}  (y_j - f(x_0^j,x_1^j, ..., x_n^j ))^2 ,$$ 
where $f(x_0^j,x_1^j, ..., x_n^j ) = \sum^n_{i=1} w_i x_i^j $ is a linear function, giving the values on the output of the neuron and depending on the weights of the input vector (instead of linear  various functions can be used, e.g. tanh, sigmoid, softmax, the general principle remains the same).
The derivative by each weight will be equal:
$$\frac{ \delta Q(\cdot)}{\delta w_i} =\frac 12 \sum^m_{j =1} \frac\delta {\delta w_i} (y_j - \sum^n_{i=1} w_i x_i^j )^2 =$$

$$ =  \sum^m_{j =1}  (y_j - \sum^n_{i=1} w_i x_i^j )(-x_i^j)$$
So, the formula for weights updating is:
$$ w_i = w_i + \mu x_i^j  \sum^m_{j =1}  (y_j - \sum^n_{i=1} w_i x_i^j )$$
Iterations are performed until the weight's changes are more than some exogenous determined parameter.
\subsection{Advantages and disadvantages}

The approach performs well for a very narrow class of problems, though it has advanced modifications like Adagrad, Adam, and Gradient Boosting which show outstanding results although require considerable computational power. 
\subsection{Computational Complexity}
Algorithm's complexity for the described case is $O(n^2)$ as on each iteration the sum of $n$ terms is computed $n$ times. The complexity is linear relative to input data size.
\section{Neurobayesian Approach}

\subsection{Description}
The bayesian approach means designing the  forecast of distribution of variables and then  updating  it based on new knowledge about an object. 
Bayesian framework  in comparison with traditional (frequentist)  one considers all data random, makes great use of Bayes theorem and instead of ML-estimates maximise posterior  probability. Though it works with any amount of data, even small,the more information about the object is obtained, the more precise the final knowledge will be (the Bayesian updating can be applied iteratively as new data becomes available).

The problem is formulated as follows: let $p(X,T,W) $ be a joint distribution of hidden variables $T$, observed variables $X$ and decision rule parameters $W$.
If training data is $ (X_{tr}, T_{tr})$, then posterior distribution of W is 
$$ p(W|T_{tr},X_{tr})  = \frac {p(T_{tr},X_{tr}|W) p(W)}{ \int p(T_{tr},X_{tr}|W) p(W) d W} $$
Analytical approach to that expression is to make use of delta-method and maximise likelihood function. But the bayesian approach proposes a less straightforward, yet fruitful step - to regularise ML estimate using prior distribution. It helps to cope with overfitting and gives a more thorough picture of hidden variables and also allows to learn from incomplete data.
But the main advantage is the possibility of the algorithm to obtain high accuracy results with big data. 
Application of bayesian approach to neural networks is based on cross-entropy error function minimisation with updating of the weights using existing data.

\subsection{Formulae} 
In practice  the optimisation procedure goes in the following way\cite{vetrov}\footnote{https://www.sdsj.ru/slides/Vetrov.pdf}[Vetrov D., 2017].
If only $X_{tr}$ is known, $T$ is a hidden variables vector,  $p(X), q(T) $ are their density functions respectively, and $W$ is to be determined, then:
$$ W_* = \arg \max p(W|X_{tr}) = \arg \max \log p(W|X_{tr}) = $$
$$ = \arg \max (\log p(X_{tr}|W) + \log p(W)) =$$ $$= \arg \max \Big( \log \int p(X_{tr}, T| W) dT + \log p(W)\Big)$$
The problem is that the first term is not concave. To cope with that a trick is introduced:
$$ \log p(X_{tr}|W) = \int  q(T)  \log p(X_{tr}|W)d T =  $$
$$ = \int q(T) \log \frac{p (X_{tr}, T |W)}{p (T|X_{tr},W)}  dT =$$
$$ =  \int q(T) \log \frac{p (X_{tr}, T |W) q(T)}{p (T|X_{tr},W) q(T)} dT =$$
$$  = \int q(T) \log \frac{p (X_{tr}, T |W)}{ q(T)}  dT + 
\int q(T) \log \frac{q(T)}{p (T|X_{tr},W) }  dT = $$
$$ \mathcal{L} (q,W) + KL(q(T)|| p(T|X_{tr},W))$$
where $ KL(q||p)$ is Kullback-Leibler divergence, which can be interpreted as distance between distribution and is always non-negative. So we can iteratively maximize  $\mathcal{L} (q,W)$ instead of original expression.
Procedure is called EM-Algorithm and consists of two steps:
\begin{enumerate}
\item \textbf{E-step}
$\mathcal{L} (q,W_{t-1}) \rightarrow \max_q$ - corresponds to KL-divergence minimization.
$$q_t (T)= \arg \min_q  KL(q(T) || p(T| X_{tr}, W_{t-1})) = p ( T| X_{tr}, W_{t-1})  $$
\item \textbf{M-step}
$\mathcal{L} (q,W_{t-1}) \rightarrow \max_W. $
$$ W_t = \arg \max_w \mathcal{L} (q,W_{t-1}) =$$
$$ = \arg \max_w \int q_t(T) \log \frac{p(X_{tr},T|W)}{q_t (T)}  dT=$$
$$ = \arg \max_w \int q_t(T) \log p(X_{tr} ,T|W) dT, $$ which is a concave function, so the maximisation is legal.
\end{enumerate}
The procedure is repeated until convergence.
\subsection{Advantages and disadvantages}
The main advantage of the approach  is  its scalability to large datasets, and less complexity in comparison with analytical optimisation solutions, which are quadratic over data size $ O(n^2)$, while EM algorithm is linear on data $ O(n)$ and weight vector's dimensions. The approach allows to combine models, trained on different data (for same consequences), and thus achieve better results. The drawback is that the approach does not take into consideration the data's structure, its intrinsic  features, which can be exploited to build more realistic a priori distributions and achieve better performance.
Since 2012, a number of studies were published, in which
a new mathematical apparatus that allows to scale Bayesian methods to big data is proposed. At the heart of it lies an interesting idea.  First, the problem of Bayesian inference (that is, the process of application of the Bayes theorem to data) was formulated as an optimization problem, and then modern techniques of stochastic optimization were applied to it, which made it possible to solve extremely large optimization problems approximately. This allowed Bayesian methods to enter the field of neural networks.Over the past 5 years, a whole class of Neurobayes models, that can solve a wider range of problems than conventional deep-seated neural networks, has been developed.  %\begin{figure}
%  \centering
%\caption{Stochastic Gradient Descent Illustration}
%  \includegraphics[scale=.3]{un5.png}
%\end{figure}
\section{Informational Neurobayesian Approach}
\subsection{Theoretical background}

Some facts from the information theory  are required for understanding this article.
At first, the number of characters that can be obtained using the alphabet consisting of n symbols is $N = m^n,$ where $n$ is the number of characters in the message. To avoid exponential dependency, it was proposed by Hartly to represent the quantity of information  as a  logarithm of number of all possible sequences:
$$ I = \log N = \log m^n = n \log m$$
Let there be $k$ characters, each has $m$ features and $p_i$ is a probability to get each feature. Shannon proposed a proportion for determining the average quantity of information in a message with random probabilities of character's values:
$$  I = - k \sum_{i=1}^m p_i \log p_i$$
The more uncertain  is the consequence, the more information can be obtained from receiving the information about the consequence.To measure the uncertainty we use entropy, which is the average quantity of information per character. 
$$ H = \frac Ik = - \sum_{i=1}^m p_i \log p_i$$
The key task in this context is to determine the amount of valuable information in a message (cause) for classification of possible consequence. For its solution it is required to calculate the amount of information, contained in $i$ feature and in the fact, that the object with this feature belongs to class $j$.

The average amount of information, contained in all features on all classes: 
 \begin{equation} I (W, M) = \sum^W_{j=1} \sum^M_{i=1} p_{ij} \log_2 \frac{p_{ij}}{p_i p_j} \end{equation}
   is precisely the average of ``individual amounts of information" in each feature about each class. 
%So, $i(x_j, y_i) = \sum \log_2 \frac{p_{ij}}{p_i p_j}$

If there are $M$ characters in the message, then the information  on belonging to class $j$   can be referenced as information's density and is expressed via Pointwise Mutual Information $$PMI (w_j, m_i) =  \log \frac{p_{ij}}{p_j p_i}= \log \frac{p_{ij}}{p_j} +\log \frac{1}{p_i} =$$ $$=\log p_{ij} - \log p_i - \log p_j$$
The expression above represents the sum  of quantities of relevant information by Harkevich and Shennon, and is interpretation of Bayes' theorem for information theory.
%It's important to note that namely the transition to informational measure allows to add them to each other for various features.
%After expressing the information from existing data, the formula will have the following form: $$i(x_j, y_i) = \sum^M_{i=1} \log_2 \frac{N_{ij}}{N_i N_j}.$$
\subsection{Foundations of System Theory of Information (STI)} 

In this paper the system generalisation of Hartley's formula is used as $ I = \log W^\varphi,$ where $W$ is a number of pure conditions of the system, $\varphi$ -- Hartley's emergence coefficient (the level of system's complexity, which consists of $W$ pure conditions). 
Lucenko [Lucenko, 2002]] took as an axiom the statement that system generalisation of Hartley's formula is \begin{equation} \varphi=\frac{ \log \sum_{z=1} ^Z C^m_W}{\log_2 W} \end{equation}
%As known from combinatorics, for case $Z=W$,  $  \sum_{z=1} ^Z C^z_W = 2^W -1$, which allows to write $I = \log (2^W-1)$. But it's just the number of maximal possible amount of information.
%The next step is consideration that in every systems there are prohibited combinations of elements from which the systems architecture depends. That's is why authors propose to name system's prohibitions as System's Informational Project.
%Nevertheless, on real data on 4 and more classes the error of approximation $I = \log (2^W-1)$ can be omitted  and Hartley's emergence coefficient can be estimated as $$ \varphi = \frac {\log \sum_{m=1} ^M C^m_W}{\log W}$$. From such representation of is is evident that the coefficient shows the level of object system's connectedness. It varies from 1 (lack of connectedness) to $\frac W{\log W} $ -- the maximal possible amount of information.
For every number of system's elements there is a maximal level of system's synergy.
According to STI, the amount of information  should be evaluated as \begin{equation} I (W, Z) = \log W + \log W^{\varphi-1}, \end{equation} in other words, it consists of classical and synergic parts. 
The system's information that we obtain from an object via STI approach is actually  information on all possible configurations of the system.
Prof. Lucenko [Lucenko, 2002] discovered a tendency for share of synergic information to raise as the number of elements increases and he propose to name it as the \textbf{Law of emergence increase}, illustrated at Figure 1.

\begin{figure*}[t]
  \centering
\caption{Emergence Increase Law}
  \includegraphics[scale=.25]{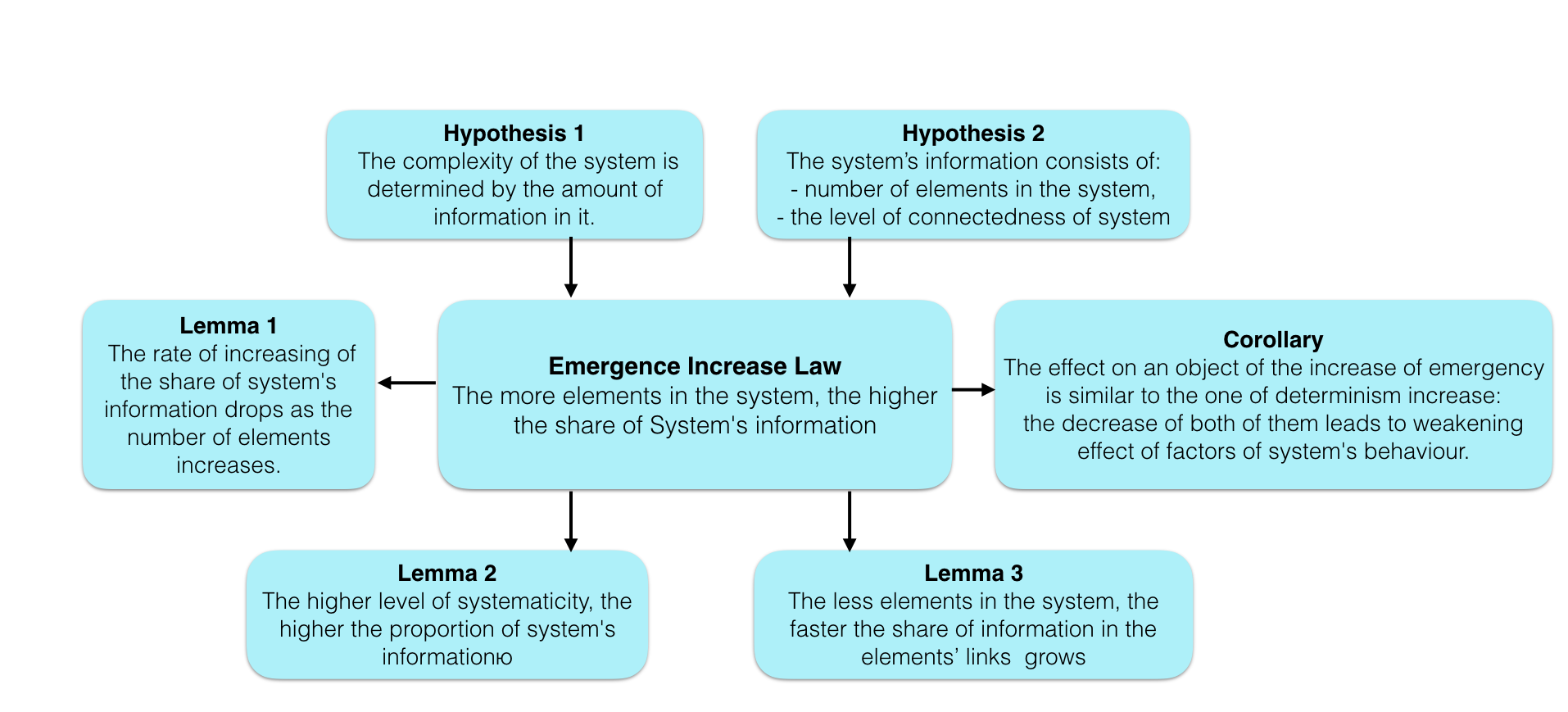}
\end{figure*}

\begin{figure*}
  \centering
\caption{Illustration of the concept of the Information Neural Bayesian approach based on the STI}
  \includegraphics[scale=.38]{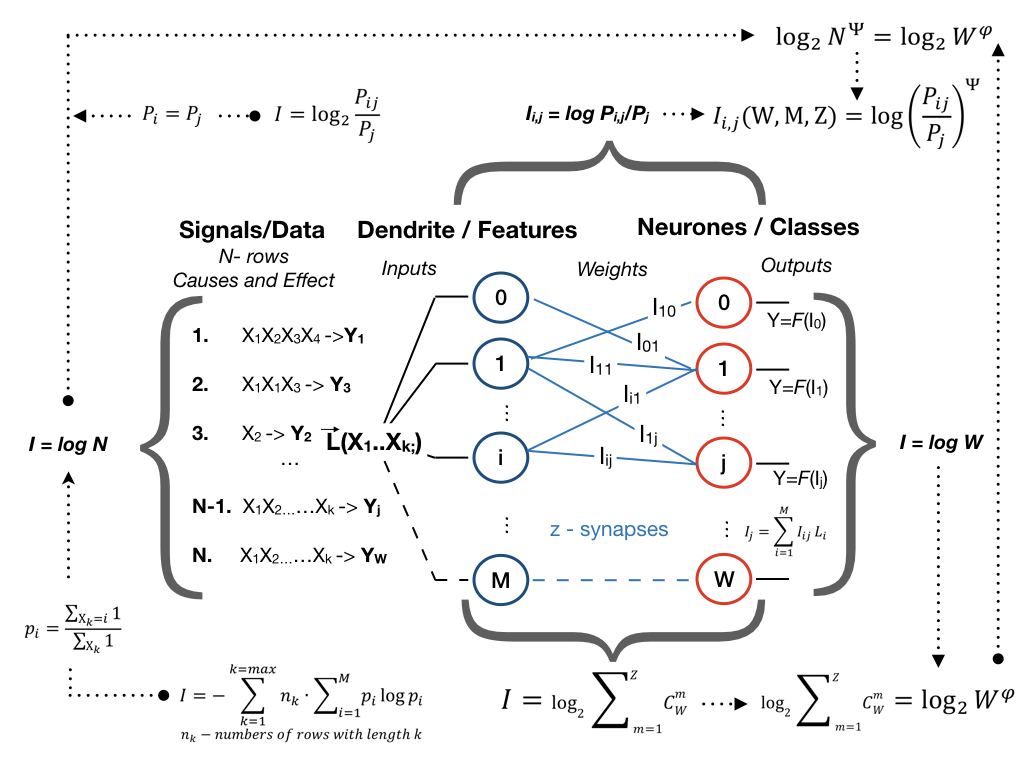}
\end{figure*}

The classical Harkevich formula is 
\begin{equation} I_{ij} = \log \frac{P_{ij}}{P_j},\end{equation} 
where $P_{ij} $ is the probability of attaining  target $J$ after getting message about feature $i$ and $P_j$ is the probability of attaining target (class) $j$ without any information. This concept does not take into account the cardinality of space of future configurations of the object, while they can be taken into consideration by involving the classical and synergic Hartley's formulas, nevertheless  it does not allow to obtain the quantity of information in bits. 
To solve that problem, we take for each feature $i, \ 1\leq i \leq M$ and class $j, \ 1\leq j \leq W$  approximate the probabilities $P_{ij}$ and $P_j$ through frequencies:
\begin{equation} P_{ij} = \frac{N_{ij}}{N_i}, \ \ P_i = \frac{N_i}{N}, \ \ P_j = \frac{N_j}{N}, \end{equation}
\begin{equation} N_i = \sum ^W _{i=1}N_{ij}; \ \  N_j= \sum^M _{j=1}N_{ij};  \ \ N = \sum ^W _{i=1}\sum ^M _{j=1} N_{ij} \end{equation}
where $N_{ij}$ is the total number of events like ``condition $j$ was obtained after feature $i$ acting'', $N_j$ - total number of features, with which the condition $j$ was obtained, $N$ - total number of of various features for all final conditions.
After placing into $ I_{ij} = \log \frac{P_{ij}}{P_j} $ values $P_{ij}$ and $P_j$ we get the expression of value of information measured as its quantity:
\begin{equation} I_{ij} (W,M) = \log_2 \frac{N_{ij}N}{N_i N_j} \end{equation} 
%т
%

As it is known, the classical Shennon formula for quantity of information for events with varied probabilities transforms into Hartley's formula with the condition that all events have the same probability, i.e. satisfy the basic property of conformance.  So the same can be applied to the Harkevich formula, which means that in marginal case it  should become Hartley's formula (marginal case means that there is unique feature for every class (object's condition) and vice versa, and all thesade classes have the same probability. In that case the quantity of information in each feature about belonging to classes is maximal and is equal to information calculated through Hartley's formula. So, in case of one-to-one correspondence of features and classes:
%$$ \forall \ i, j \ N_{ij} = N_{i} = N_{j} = 1 $$
%Harkevich's formula has the following form:

General formula of Lutsenko formula and his emergence coefficient $\psi$ are:
\begin{equation} I_{ij} = \log_2 N^{\psi} = \log_2 W^\varphi, \end{equation}

\begin{equation} \psi = \frac{\log_2 W^\varphi}{\log_2 N} \end{equation}
Taking it into account,  authors suggest  to add an emergence coefficient into the modified Harkevich formula. The resulting expression is named Lutsenko emergence information quantity formula: 
\begin{equation} I_{ij} (W,M,Z)= \log_2 \Big(\frac{N_{ij}N}{N_i N_j}\Big)^\psi,\end{equation}
where $\psi$ is Harkevich's emergence coefficient, which defines the degree of determination of the object with the system's organisation level $\varphi$, having $W$ pure conditions and $M$ features, on which the final condition is dependent, and Z is the maximal complexity calculated for each factor separately.  $N$ observations  on system's behaviour were made. That generalisation can be performed in the following way:
%Finally, taking into account Hartley's emergence coefficient, we get it's advanced modification, Lucencko emergence coefficient:
%$$  \psi = \frac{\log_2 W^{\frac{\log_2 \sum^{M}_{m=1} C^m_M}{\log_2 W}}}{\log_2 N} $$
%Harkevich's emergence coefficient's meaning becomes more clear after the fact that if $W=N$m then it equals 1, as in that case features uniquely determine object's condition.
%If $N>W$, which means that features don't solely determine the condition of the object, Harkevich emergence condition varies from 0 to 1 and determines the degree of determination of the system: 
%\begin{enumerate}
%\item $\psi=1$  corresponds to absolutely determined system,  behaviour of which is depends on minimal number of features:
%\item$\psi=0$ corresponds to totally random system.
%\item $0<\psi<1$ corresponds to  system, where there are more features than classes and no one of them has the key role for determining the class.

\begin{equation} I_{ij} = \log_2 \Big(\frac{N_{ij}N}{N_i N_j}\Big)^\psi = \log_2 \Big(\frac{N_{ij}N}{N_i N_j}\Big)^{\frac{\log_2 W^\varphi}{\log_2 N}} = \end{equation}
\begin{equation}=\frac{\log_2 W^\varphi}{\log_2 N} \Big( \log_2 \Big(\frac{N_{ij}}{N_i N_j}\Big)  +\log_2 N \Big) = \end{equation} 
\begin{equation}=\log_2 \Big(\frac{N_{ij}}{N_i N_j}\Big)^{\frac{\log_2 W^\varphi}{\log_2 N}}  + \log_2 W^{\varphi}
\end{equation}

The same calculations for the continuous case:

$$ I = \log_2 \int^{Z}_{z=q} \Big( \frac{G(W)}{G(z)\cdot G(W-z) }\Big)  = $$
\begin{equation} = \log_2 \left \{ G (W) \cdot \int ^{Z}_{z=1} \frac{\,d z}{G(z)\cdot G(W-z)}\right \} =\end{equation}
$$ = \log_2 G(W) + \log_2 \int^{Z}_{z=1} \frac{\,d z}{G(z)\cdot G(W-z)} $$
And finally:
\begin{equation} I = \log_2 G(W) + \log_2 \int^Z_{z=1}\frac{\,d z}{G(z)\cdot G(W-z)} = I(W) + I(W, Z) \end{equation}

Generalised Kharkevich \cite{khar2} formula also satisfies correspondence principle, i.e, transforms into Hartley's formula:
$\log \frac{P_{ij}}{P_j} = \log \frac{N_{ij}N}{N_i N_j}=\log N.$
In the marginal case, when  every class (condition of the object) corresponds  only one feature, and for every feature - one class,  $\forall \ \  N_ij = N_i = N_i =1 $ these classes and features have the same probabilities.

%It's worth to mention that the coefficient is derived for each feature separately, e.g.:
%$$ I_{ij}=\log_2 N^{\psi} = \log_2 W^{\psi}$$, where  $N$ is the number of pairs class-feature.
%
%$$ I_{ij} =  \log_2 \Big(\frac{N_{ij}N}{N_i N_j}\Big)^\psi   = 
%\log_2 \Big(\frac{N_{ij}N}{N_i N_j}\Big)^{\frac{\log_2 W^{\psi}}{\log_2 N}} =$$
%$$ = \log_2 W^{\psi}{\log_2 N} \Big( \log_2 N + \log_2 \Big( \frac{N_{ij}}{N_i N_j}\Big) \Big) = $$
%$$ = \log_2 W^{\psi} +\log_2 \Big( \frac{N_{ij}}{N_i N_j}\Big) ^{\frac{\log_2 W^{\psi}}{\log_2 N}}$$
%As there are $W$ classes and complexity $Z$:
% $$ \varphi = \frac{\log_2  \sum_z^Z C^z_W}{\log_2 W} = \frac{log_2  \sum_m C^{m=1}_W}{\log_2 W}$$
% $$ \sum_m C^m_W  = 2^W -1, \ Z= W$$% \forall m>4$$

\subsection{System Emergence Coefficient}
General formula of Lutsenko for Harkevich emergence coefficient:
%$$ I_{ij} = \log_2 N^{\psi} = \log_2 W^\varphi, $$
\begin{equation}
 \psi = \frac{\log_2 W^\varphi}{\log_2 N}
\end{equation}

% emergence coefficient's meaning becomes more clear after the fact that if $W=N$ then it equals 1, as in that case features uniquely determine object's condition.
where $\varphi = \frac{\log_2 \sum^Z_{m=1} C^m_W}{\log_2 W} $.\\

$\psi$  varies from 0 to 1 and determines the degree of determination of the system: 
\begin{enumerate}
\item $\psi=1$  corresponds to absolutely determined system,  behaviour of which depends on the minimal number of features.
\item$\psi=0$ corresponds to totally random system.
\item $0<\psi<1$ corresponds to  system, where there are more features than classes and none of them plays the key role in determining the class. 
\end{enumerate}
%As was stated before,
 Professor Lutsenko\cite{luc2} proposed coefficients representing the degree of determination of the possible state of the object with a set of features at the system organization level  $\varphi$. Nevertheless, as it seems hard to evaluate, it was decided to omit that concept and admit the coefficient $\varphi$ to be equal 1 (minimum level of complexity).
 \begin{equation} \psi = \frac{\log_2 W}{\log_2 N}
 \end{equation}

 Straightforward way to calculate it is to consider all possible of combinations and classes, and for each one to evaluate the amount of information. Evidently, it will require enormous calculations.
% \begin{equation} \varphi = \frac{log_2  \sum_{m=1}^Z C^{W}_m}{\log_2 W}\end{equation}
% 
% \begin{equation}\psi = \frac{\sum_{m=1}^Z C_W^m}{\log N}  \end{equation}
 \begin{equation}\psi = \frac{\log_2 W^{\varphi}}{\log_2 N} =\frac{\log_2 W^{\frac{log_2  \sum_{m=1}^Z C^{W}_m}{\log_2 W}}}{\log_2 N} \end{equation}
The Lucenko-Artemov formula for  emergence coefficient of system:
% $\varphi = \frac{\log_2 \sum^Z_{m=1} C^m_W}{\log_2 W} $:
\begin{equation}
\psi = \frac{log_2 \sum^Z_{m=1} C_W^m}{\log_2 N}
\end{equation}
The value obtained is the emergence coefficient of the system for the real level of complexity. The optimal solution is to consider only relevant combinations of feature-classes, dividing them into groups by complexity, up to which the combinations will be evaluated.
There are two approaches for calculation the complexity Z. The first one for each feature takes the number of classes the feature corresponds to  (gives positive amount of information). The number of classes represents the complexity, up to which the combinations of classes will be calculated. 
The second method is to divide all features and classes into groups with the same number of features as well as classes. The idea is to take all classes to whom the feature corresponds and select other features which correspond to the same classes. The key point is that it helps to build one-to-one correspondence between features and classes within that group. 
Finally we obtain limited number of features $i$ or groups of
features $i \in g(\sum_g i= |M|$, where $M$ is the number of features), which are acting collectively $N_g$ times in the same number of classes $W_g$. That means  for given features maximal complexity is $Z_g = W_g$. So $ \sum ^{Z_g}_{m=1} W_g = 2^{Z_g}-1| Z_g = W_g, g \in M$, and we get the final expression for the coefficient of emergence for each group:
 \begin{equation}
  \psi_g= \frac{log_2 (2^{W_g} -1)}{\log_2 N_g}  \end{equation}
 
The coefficient is normalised by the following rule:
\begin{equation}
    \psi_g  \ = \ 
    \begin{cases}
      1, & \text{if} \ W_g \textgreater \log_2 2N \\
      \frac{log_2 (2^{W_g}-1)}{\log_2 N}, &\text{otherwise}
    \end{cases}
\end{equation}
Formula 21 is the Artemov-Lucenko emergence coefficient, for the system of features, aggregated by the maximal level of of system's complexity.
% For all features the ranges of classes can are distributed in different ways. So all of them can be divided by belonging to the sets of possible conditions (classes) $w_i$.There are two ways to carry out that procedure. The first is to divide all classes into groups which can be obtained from some combination of features. All considered combinations should be selected before training. The second one is to take all features with the same ranges of classes, and so all features will be grouped by classes.
% $$ \varphi  = \frac{\log_2 (2^W -1)}{\log_2 W} $$
% $$  \psi  = \frac{\log_2 W^{\varphi}}{\log_2 N}$$
% $$ \log_2 W^\varphi = \log_2 W ^{\frac{2^W -1}{\log_2 W}} =$$ $$= \frac{2^W -1}{\log_2 W}  \log_2 W = 2^W -1$$
%$$ \psi = \frac{\log_2 W^{\varphi}}{\log_2 N} \approx \frac{2^W -1}{ \log_2 N},  \ \forall m > 4$$
%$$ \psi_i = \frac{2^{Z_i}-1}{\log_2 N^{z_i}} $$
%
%The last formula is author's modification for the Lutsenko formula, named as Lutsenko-Artemov coefficient. It allows to take into consideration the emergence of features, when they show up together. 
%
%

\subsection{Application to neural networks}
The essence of the proposed approach shows its best  application to neural networks. The main idea is that the quantity of information, calculated for all feature-class pairs can be expressed in network's weights, so they gain a meaning instead of being simply empirical parametrs. It explains the given name "Informational Neurobayesian" for the approach. Every weight represents  quantity of information - the  object with the feature $i$ activates given neuron $j$:
\begin{equation} I_{ij} = \psi_{i} \  \log{\frac{P_{ij}}{P_j}} ,  \end{equation} %\ \  b = I_0 - bias  
where $\psi_i$ - coefficient of emergence of the system for feature  $i$.

\begin{figure}
\centering
\caption{Summation Process Architecture}

\includegraphics[scale=.3]{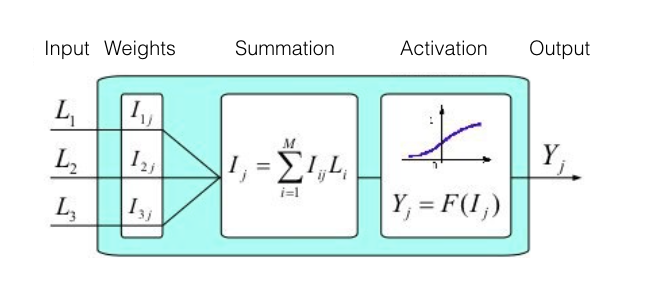}
\end{figure}
In the scheme above the $L_i$ is $i$-th object's feature. $\vv{L}\{X_1, ...,X_k\} $--  a vector of object features:

\begin{equation}
    L_i  \ = \ 
    \begin{cases}
      1, & \text{if} \  X_k \geq \epsilon  \\
      0, &\text{otherwise}
    \end{cases}
  \end{equation}

For every object with a given set of features the neural network  chooses over all classes the one with maximal value of the activation function, represented as an activated sum by all features. The process is illustrated in Figure 3.
\begin{equation}
  S_{j}=\sum^M_{j=1} I_{ij}\cdot L_{i}+\text{bias}, \ \text{bias} = I_0
\end{equation}
Bias is a parameter for activation of class. It allows to solve two problems: 1. To set a minimum threshold of neuron activation. 2. Allows integration with other models.
\begin{equation}
 j^*= \arg \max_j f(S_j)
  \end{equation}
\begin{equation}
y = f(S_{j*})
  \end{equation}

The Y is a chosen class for the given features, $f$ is activation function.
The whole workflow is presented in Figure 2.

\subsubsection{Generalised EM-Algorithm for INA }
EM-algorithm consists of two steps iteration repetions. On the E-step expected hidden variables values are calculated using current estimate of parameters. On the M-step likeli- hood is maximised and new parameters vectors estimate is being obtained based on their current values and hidden variables vector. \\

\textbf{Step 0}: Calculate a frequency matrix for input data. 

Unique features $M$ and classes $W$ are determined in training data and their frequency table which reflects the distribution of parameters $N_{ij}$ - number of features $i \in M$ for each class $j \in W$. Also $Z_g$ parameter (number of classes, in which a features group $g$ is active)  is determined during this step.   \\

\begin{figure}
\resizebox{\textwidth}{!}{

  \centering
\caption{Frequency matrix}
  \includegraphics[scale=.4]{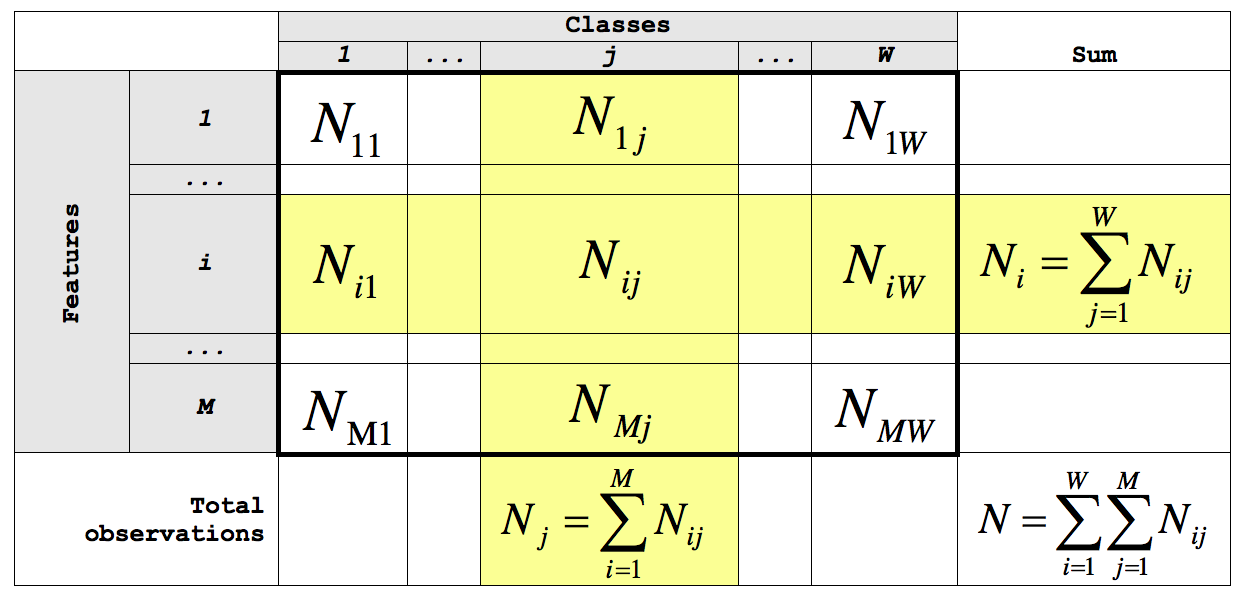}
  }
\end{figure}

\textbf{Step E}: Creating informational neurobayesian model. \\
During the E-step expected value of hidden variables  $x_i$ by current estimation of parameter vector $N_{ij}$ is evaluated. The data is represented in Figure 4. and the output is organised in the same way, as presented in Figure 5.
\begin{equation}
I_{i,j}  (W, M, Z) = \log \big( \frac{N_{i,j}N}{N_i N_j}\big)^{\psi_g}
\end{equation}
\begin{equation}
\psi_{g}=\frac{log_2 (2^{W_g}-1)}{\log N_g}
\end{equation}
\begin{equation}
I_{i,j}  (W, M, Z) \rightarrow \max_Z 
\end{equation}
% \usepackage{multirow}
\begin{figure}
\resizebox{\textwidth}{!}{

  \centering
\caption{Informational neurobayesian model}
  \includegraphics[scale=.4]{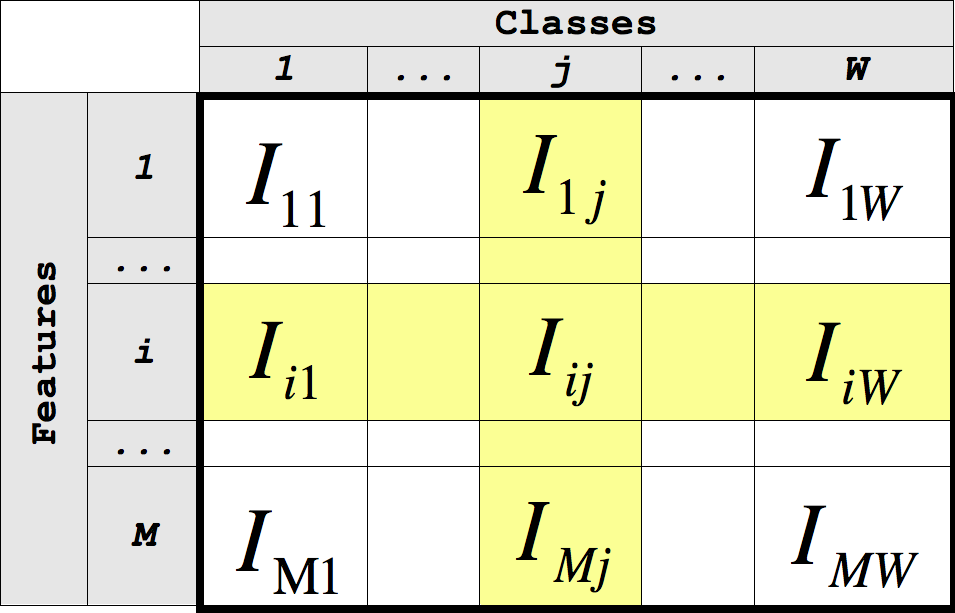}
  }
\end{figure}
%\begin{table}[]
%\centering
%\caption{Information matrix}
%\label{my-label}
%\begin{tabular}{|c|l|l|l|l|l|l|}
%\hline
%\multicolumn{2}{|l|}{\multirow{2}{*}{}} & \multicolumn{5}{l|}{Classes}                  \\ \cline{3-7} 
%\multicolumn{2}{|l|}{}                  & 1         & ... & j         & ... & $W$       \\ \hline
%\multirow{5}{*}{Features}     & 1       & $I_{1,1}$ &     & $I_{1,j}$ &     & $I_{1,W}$ \\ \cline{2-7} 
%                              & ...     &           &     &           &     &           \\ \cline{2-7} 
%                              & i       & $I_{i,1}$ &     & $I_{ij}$  &     & $I_{i,W}$ \\ \cline{2-7} 
%                              & ...     &           &     &           &     &           \\ \cline{2-7} 
%                              & $M$     & $I_{M,1}$ &     & $I_{M,j}$ &     & $I_{M,W}$ \\ \hline
%\end{tabular}
%}
%\end{table}

\textbf{Step M}: Model maximisation by Van Rijsbergen's effectiveness  measure  (F-measure). 
\begin{enumerate}
\item Calculate  $E(F_{micro}):$

 Calculate average  micro $F$-measure for each observed class $j_l \in W$ based on training set, comparing the result with $j_t$ - test dataset for object's features vector with the same set of features, $t$ stands for test data, $l$ -  data the model has been trained on («model has learnt»).
 \begin{equation}
E (F_{\text{micro}})  = \frac{\sum_k F_{j_t}}{k} 
\end{equation}
\begin{equation}
 F_{j_t}=(1+\beta)^2 \frac{\text{ precision}_{j_t} \times \text{ recall}_{j_t} }{\beta^2\text{ precision}_{j_t}  + \text{ recall}_{j_t}  } 
 \end{equation}
  $$0<\beta<1,$$ 

\begin{equation}
%S_{j}=\sum^M_{i=1} I_{ij}* L_{i}
 j_t = \arg \max F(S_{j_t})
\end{equation}

%\begin{equation}
%j_i = \arg \max F(S_j)
%\end{equation}

%$$F1_{j_t} = (1+\beta)^2 \frac{\text{average precision} \cross \text{average recall}}{\beta^2\text{average precision} + \text{average recall} } $$

\item For each $j_t \in W,$ where $F_{j_t} <  \ E(F_{j_t})$, i.e. $j_l \ne \arg \max F(S_{j_t})$ or $S_{j_t}<S_{j_l}.$ \\
2.1 Determine $j^*_t$ and $j^*_l$ - correct class for the set of features $L_i$, $i \in M$.\\
2.2 Exclude all objects $L_i$ with the same set of features, but $j^*_t  \ne j^*_l$  classes
\\
2.3 For remaining objects $L_i$ change quantity of information  for each feature, acting in given class $I_{j_l } \rightarrow I_{j^*_l}$, so that $S_{j_l } \rightarrow \max S_{j^*_l}$
$$ j_l = \arg \max F(S_{j^*_l}) $$
\\
2.4 Calculate  \begin{equation}
E (F^*_{\text{micro}})  = \frac{\sum_k F_{j_{t*}}}{k} 
\end{equation}
The updated model is checked for improvement on wrongly classified objects. \\
2.5 Continue until $E(F^*_{\text{micro}}) \ \textgreater \
E(F_{\text{micro}})$
\end{enumerate}
\begin{table*}[t]
\centering
\caption{Experiments Results}
\label{my-label}
\resizebox{\textwidth}{!}{

\begin{tabular}{|l|l|l|l|l|ll}
\cline{1-6}
\bf{Model}                    & \bf{MNIST}              & \bf{House Prices} & \bf{Emotions in text}      & \bf{Russian words}                & \multicolumn{1}{l|}{\bf{Russian words}}               &  \\ \cline{1-6}
\multirow{2}{*}{\bf{Problem}} & Hand-written       & Real Estate  & Determining  emotions & \multicolumn{2}{l|}{\multirow{2}{*}{Word recognition by letters and bigramms.}} &  \\
                         & digits recognition & Valuation    & (by Ecman), in text.  & \multicolumn{2}{l|}{}                                                           &  \\ \cline{1-6}
\bf{Features}                 & 2 815              & 809          & 19 121                & 1042                         & \multicolumn{1}{l|}{1042}                        &  \\ \cline{1-6}
\bf{Classes}                  & 11                 & 658          & 8                     & 5 099 300                    & \multicolumn{1}{l|}{5 099 300}                   &  \\ \cline{1-6}
 \bf{Parameters (weights)}                  & 30 965             & 532 322      & 152 968               & 6 313 470 600                & \multicolumn{1}{l|}{6 313 470 600}               &  \\ \cline{1-6}
\bf{Model Size (Mb)}          & -                  & 1,89         &                       & 634,94                       & \multicolumn{1}{l|}{634,94}                      &  \\ \cline{1-6}
\bf{Training data size (Mb)}  & 74,9               & 0,53         & 2,85                  & 505,43                       & \multicolumn{1}{l|}{505,43}                      &  \\ \cline{1-6}
\bf{Training data, rows}      & 42 000             & 1460         & 8 682                 & 5 099 300                    & \multicolumn{1}{l|}{5 099 301}                   &  \\ \cline{1-6}
\bf{Learning Type}            & E  (1 epoch)                & E     (1 epoch)       & E   (1 epoch)                  & E     (1 epoch)                       & \multicolumn{1}{l|}{EM}                          &  \\ \cline{1-6}
\bf{Brain2 v.}\footnote{Framework's version, developed by project's team, including paper's author}               & 1.0                & 2.0          & 2.0                   & 3.0                          & \multicolumn{1}{l|}{3.0}                         &  \\ \cline{1-6}
\bf{Result}                   & CA \footnote{This competition is evaluated on the categorisation accuracy of your predictions (the percentage of images you get correct).}=0,81          & RMSE=0,42    & F1 = 0,81             & F1=0,97                      & \multicolumn{1}{l|}{F1=0.98\footnote{Perplexity of 1.26 per letter and bigrams (feature)}}                     &  \\ \cline{1-6}
\bf{Time} \footnote{ Tests were run on the machine with specifications: Intel® Xeon®, E5-1650 v3 Hexa-Core Haswell  – 6 cores, 128 GB ECC RAM, 2 x 240 GB 6 Gb/s SATA SSD , 2X2Tb SATA3. GPUs were not used for training.
}                  & 20 min             & 2 min 33 sec & 6 min 4 sec           & 2 h, 35 min                  & \multicolumn{1}{l|}{2 h, 49 min}                 &  \\ \cline{1-6}
\end{tabular}}
\end{table*}
\begin{table*}[t]
\centering
\caption{Weights calculating methods comparison}
\label{my-label}
\resizebox{\textwidth}{!}{%
\begin{tabular}{|l|l|l|l|}
\hline
 & \begin{tabular}[c]{@{}l@{}}\bf{Gradient Descent applied to neural networks}\end{tabular} & \begin{tabular}[c]{@{}l@{}}\bf{Neurobayesian}\\ \bf{Approach}\end{tabular} & \begin{tabular}[c]{@{}l@{}}\bf{Informational} \\ \bf{Neurobayesian} \\ \bf{Approach}\end{tabular} \\ \hline
\bf{Optimisation framework} & Backpropagation & \begin{tabular}[c]{@{}l@{}}Stochastic Gradient Descent \\ with reparameterization trick.\end{tabular} & \begin{tabular}[c]{@{}l@{}}EM - algorithm applied to quantity \\ of emergent information\end{tabular} \\ \hline
\bf{Weights Interpretation} & \begin{tabular}[c]{@{}l@{}}Weights are \\ just numbers\end{tabular} & \begin{tabular}[c]{@{}l@{}}Weights represent\\  probabilities\end{tabular} & \begin{tabular}[c]{@{}l@{}}Weights represent \\  quantity of information\end{tabular} \\ \hline
\bf{Complexity} & $O(n^2)$ & $O(n)$ & \begin{tabular}[c]{@{}l@{}}$O(\log n)$ - E-step\\ $O(n\log n)$ - M-step\end{tabular} \\ \hline
\bf{Advantages} & Wide range for applications & Decreased computation complexity & \begin{tabular}[c]{@{}l@{}}1.The E-step gives satisfactory results. (no need to do M-step)\\ 2. Arithmetical operations can be applied to weights \\ as they represent the quantity of information. \\ 3.Applicable for training complex models just on CPU. \end{tabular} \\ \hline
\end{tabular}%
}
\end{table*}

\section{Experiments}
For conducting the tests we used Brain2 - a software platform implementing Informational Neurobayesian Approach. The biggest built neuromodel is aimed at word recognition by letter and bigrams in Russian language for 100 000 words – it consists of 1042 features and 5 099 300 classes, which makes the total number of parameters 6.3 bln (weight coefficients). An important factor here is that it took only 2 h 35 min to build the model on an average server without GPU (205 GFlops, 8 cores CPU). In comparison, the Google Brain biggest neural network consists of 137 billion parameters, but if we consider the needed resources – only for 4.3 bln parameters it took them 47 hours with 16-32 Tesla K40 GPUs (1220 GFlops) \cite{shaz}[Shazeer N, 2017]. On other basic machine learning tasks such as MNIST and House Prices our models have shown an acceptable result – 0.81 on MNIST (Kaggle task) with the best 1.0; and 0.42 on House Prices with the best one at 0.06628 (For one epoch). It is important to note that we did not set the task to achieve the best possible result and have trained the model as is. Instead, we have concentrated on a more difficult problem, connected with natural text – reconstruction of words with errors and finding answers on questions, where we have reached accuracy close to 1. Experiment results are presented in  Table 1.

% \usepackage{multirow}
% \usepackage{graphicx}
% \usepackage{graphicx}

\section{Conclusion}
In the paper the INA method aimed to cope with complex classification problems more intelligently than currently used analytical and numerical methods was described.  Its ability to be applied to neural networks and provided experiments showed opportunities for the suggested method to simplify training process in various classification problems and achieve  at least the same performance with less computational costs. The results of the experiments are presented in Table 1. And the summary of comparison with other method is presented in Table 2.

The additive properties of the amount of information allow not only  to explain the correctness of  addition of weights while choosing a class within a single model (layer), but also to link different models (layers),  transferring values ​​of the amount of information from  one   input to another. In addition, every information neuromodel can be trained independently. Eventually it makes possible to facilitate the construction of super-neural networks - multilevel (and multi-layer) neural networks, where each input of the neuron is a neural network of a lower order (from the position of information amount).
 The proposed information approach makes it possible to obtain an acceptable level of quality of the model (F1 $>$ 0.6, for 1 epoch), even on relatively small data. And given the low complexity $(O(n))$ of the learning algorithm , the INA opens new horizons to the training of superneuromodels on conventional (not super) computers.

 In our opinion, the INA combines in itself  a general philosophical view presented by such cognitive scientists as D. Chalmers\cite{chal}\footnote{[Chalmers,1996]} (as each layer represents a level of cognition, influencing a higher cognition level), V. Nalimov (in the works on the  \cite{nal}\href{https://www.tib.eu/en/search/id/BLCP%3ACN026838838/V-V-Nalimov-s-probabilistic-model-of-language-and/ }{probabilistic model of language}  [Nalimov, 1990]) 
 and the pragmatic concept of meaning - the amount of information in the cause for a given consequence required  for making a managerial decision \footnote{[Kharkevich,2009]} \cite{kha} (Ashby\cite{ash} \footnote{[Ashby,1961]}, Wiener, U. Hubbard).

\end{document}